\documentclass[a4paper,fleqn]{cas-dc}
\usepackage[authoryear]{natbib}
\usepackage{graphicx}
\usepackage{enumitem}
\usepackage{caption}
\usepackage{amsfonts}
\usepackage{amsmath}
\usepackage{booktabs} 
\usepackage{multirow} 
\usepackage{hyperref}
\usepackage{tabularx}
\usepackage{array}

\begin{document}
\shorttitle{DAPNet for Network State Classification}
\shortauthors{Xuelong Wang et~al.}

\title [mode = title]{Dynamic Adaptive Parsing of Temporal and Cross-Variable Patterns for Network State Classification}

\author[1]{Yuan Gao}[style=chinese, 
						orcid=0009-0002-7561-9847]
\cormark[1]
\ead{gy@besti.edu.cn}
\credit{Conceptualization of this study, Methodology, Software}

\author[1]{Xuelong Wang}[style=chinese,
						orcid=0009-0007-7134-1039]
\ead{20243806@mail.besti.edu.cn}
\credit{Data curation, Writing - Original draft preparation}

\author[1]{Zhenguo Dong}[style=chinese]
\ead{20243808@mail.besti.edu.cn}

\author[2]{Yong Zhang}[style=chinese,
						orcid=0000-0001-6650-6790]
\ead{zhangyong2010@bjut.edu.cn}

\affiliation[1]{organization={Beijing Electronic Science and Technology Institute},
	postcode={100070}, 
	city={Beijing},
	country={China}}

\affiliation[2]{organization={Beijing University of Technology},
	postcode={100124}, 
	city={Beijing},
	country={China}}

\cortext[cor1]{Corresponding author}

\begin{abstract}
Effective network state classification is a primary task for ensuring network security and optimizing performance. Existing deep learning models have shown considerable progress in this area. Some methods excel at analyzing the complex temporal periodicities found in traffic data, while graph-based approaches are adept at modeling the dynamic dependencies between different variables. However, a key trade-off remains, as these methods struggle to capture both characteristics simultaneously. Models focused on temporal patterns often overlook crucial variable dependencies, whereas those centered on dependencies may fail to capture fine-grained temporal details. To address this trade-off, we introduce DAPNet, a framework based on a Mixture-of-Experts architecture. DAPNet integrates three specialized networks for periodic analysis, dynamic cross-variable correlation modeling, and hybrid temporal feature extraction. A learnable gating network dynamically assigns weights to experts based on the input sample and computes a weighted fusion of their outputs. Furthermore, a hybrid regularization loss function ensures stable training and addresses the common issue of class imbalance. Extensive experiments on two large-scale network intrusion detection datasets (CICIDS2017/2018) validate DAPNet's higher accuracy for its target application. The generalizability of the architectural design is evaluated across ten public UEA benchmark datasets, positioning DAPNet as a specialized framework for network state classification.
\end{abstract}

%

\begin{keywords}
Network Classification \sep Time Series \sep Mixture of Experts \sep Dynamic Cross-Variable Correlation \sep Periodicity Analysis
\end{keywords}

\maketitle

\section{Introduction}
In the contemporary digital era, computer networks serve as the foundational infrastructure for global commerce, communication, and critical services. The proliferation of the Internet of Things (IoT), cloud computing, and diverse online applications has triggered an exponential surge in the volume and complexity of network traffic, which is naturally represented as high-dimensional Multivariate Time Series (MTS). The ability to accurately classify network states in real-time is essential for operational stability and security, moving beyond a purely academic pursuit. This capability is a key component of proactive cybersecurity, enabling the detection of sophisticated threats such as large-scale Distributed Denial of Service (DDoS) campaigns, stealthy reconnaissance scans, and internal network infiltrations that manifest as anomalous, coordinated behaviors across multiple system variables. As cyber threats, including those driven by artificial intelligence, grow in sophistication, this trend of increasing threat sophistication necessitates the development of advanced classification frameworks. 

\begin{figure}
    \centering
    \includegraphics[width=1\linewidth]{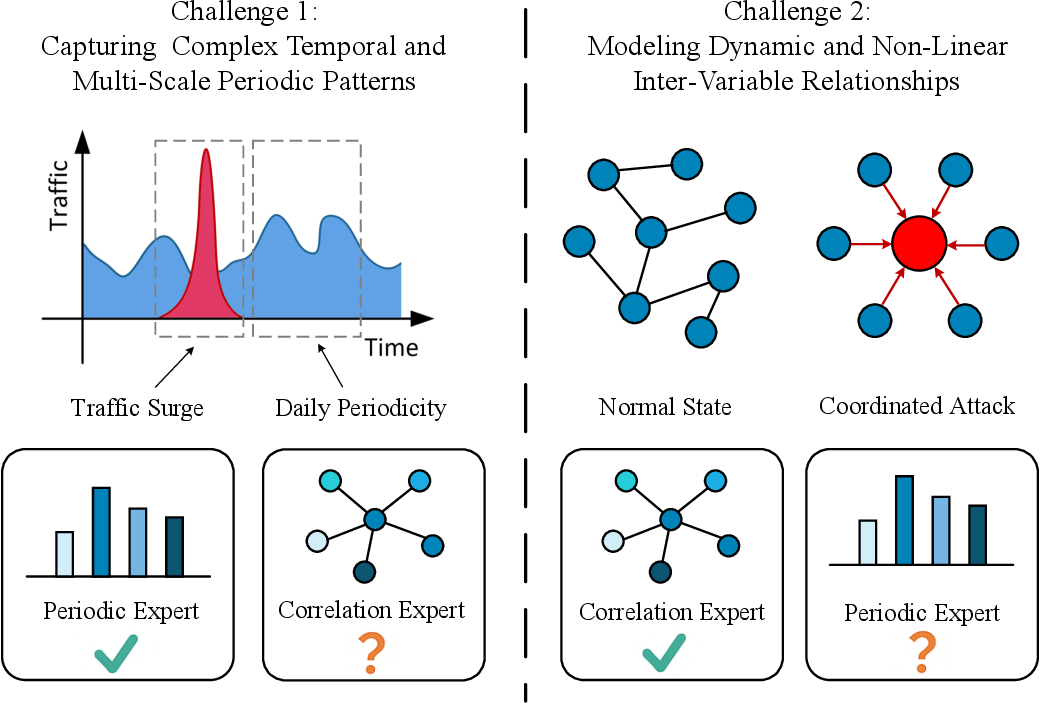}
    \caption{The Challenge of Heterogeneous Patterns in Network State Classification.}
    \label{fig:1}
\end{figure}

In response, the field has increasingly turned to deep learning, building upon a long history of applying machine learning to intrusion detection \citep{45}, as deep models have demonstrated a superior capacity to automatically learn hierarchical feature representations from raw, high-dimensional network traffic data \citep{19}, including in challenging scenarios like encrypted traffic \citep{20}. This paradigm shift has led to the development of specialized architectures, each tailored to extract distinct types of patterns.

As shown in \autoref{fig:1}, the analysis of network MTS data is complicated by the presence of two fundamental and orthogonal pattern types, which presents a distinct challenge in model design. The first pattern type involves complex temporal dependencies, where network activity exhibits rich, multiscale periodicities that are superimposed and intertwined, ranging from microsecond-level packet bursts to broader diurnal and weekly usage cycles \citep{6}. The second pattern type concerns the dynamic, non-linear inter-variable relationships \citep{8}. Contrary to the common modeling assumption of independence, network features are interconnected. These relationships contain discriminative information. A prime example is the transient correlation topology that emerges during a DDoS attack, where thousands of source IP features become highly correlated with a single destination port feature. This duality presents a significant modeling challenge, as deep learning architectures are often specialized to capture one of these pattern types with high fidelity, which can create inherent trade-offs in their ability to model the other. The core challenges can be summarized as follows:

\begin{itemize}[leftmargin=*]
\item Capturing the complex temporal patterns with multiscale periodicities, ranging from microsecond-level packet bursts to broader diurnal and weekly usage cycles.
\item Modeling the dynamic, non-linear relationships between network features, which can form transient graph-like structures that contain discriminative information relevant to classification.
\end{itemize}

To contend with these challenges, distinct research thrusts have emerged, each developing specialized architectures that excel at one of these analytical dimensions. Periodicity-centric models, such as TimesNet, leverage frequency domain analysis and 2D convolutions to achieve competitive performance in capturing intricate, multiscale temporal patterns, but their core assumption of channel independence prevents the modeling of cross-variable correlations. Conversely, graph-based models like TodyNet are explicitly designed to model these inter-variable dependencies using Graph Neural Networks, making them adept at identifying collaborative behaviors, yet this focus can come at the cost of sacrificing fine-grained temporal precision. While the Mixture-of-Experts (MoE) paradigm has emerged as a promising architecture capable of combining multiple specialized sub-models, its current applications to network state classification have yet to resolve this fundamental challenge.

Based on this observation, we design a framework called the Dynamic Adaptive Parsing Network (DAPNet), which is designed to integrate these distinct modeling paradigms. Rather than assuming that any single model is universally dominant, we posit that different types of network traffic may contain key information in different dimensions. The core idea of DAPNet is to adapt to the characteristics of each input sample. Our contributions can be summarised as follows:

\begin{itemize}[leftmargin=*]
\item We design a framework of experts specializing in periodicity analysis, dynamic cross-variable correlation modeling, and hybrid temporal feature extraction to address the multifaceted nature of network state classification. 
\item A hybrid regularisation loss function is introduced, combining Focal Loss with a load-balancing loss, which collectively enhance model performance and training stability.
\item Extensive experiments on large-scale network security datasets and diverse public benchmarks show that DAPNet achieves higher performance metrics than the evaluated baselines, the results suggest its utility as a framework for network state classification with notable generalization capabilities.
\end{itemize}

\section{Related Work}
The evolution of network state classification has progressed from early machine learning techniques \citep{40} toward deep learning, initially employing models such as Deep Belief Networks \citep{1} and hybrid CNN-LSTM \citep{2} architectures to capture temporal dynamics. However, as the complexity of modern network traffic and the sophistication of threats grew, the field saw the emergence of highly specialized architectural paradigms \citep{38}, each tailored to extract distinct types of patterns from multivariate time series data.

One prominent research thrust has focused on capturing intricate temporal dynamics with high fidelity. Models adopting a "channel-independent" strategy, such as PatchTST \citep{4}, excel at processing the unique sequential evolution of each variable in isolation. Concurrently, frequency-domain models like TimesNet \citep{6} and MPTSNet \citep{7} have proven highly effective at identifying multi-scale periodicities by transforming time series into 2D representations for analysis. Similarly, other architectures have found success by embedding series decomposition directly into the model as a core principle, separating trend and seasonal components to simplify the learning task \citep{47}. This specialization is not unique to frequency-domain models, such as applying Vision Transformers to time series as images \citep{23} or relying on pure MLP-based structures \citep{48}, also tend to prioritize one dimension over the other. While powerful for tasks where such temporal patterns are dominant, their architectural design, by focusing on individual channel characteristics, does not explicitly model the dynamic cross-variable correlations that can define coordinated network events like DDoS attacks.

A complementary research thrust prioritizes the modeling of these cross-variable relationships. Models like iTransformer \citep{5} invert the channel-independent approach, while graph-based models such as TodyNet \citep{8} and WinGNN \citep{9} leverage Graph Neural Networks (GNNs) to dynamically model the relational structure between variables, a paradigm whose rapid evolution has been well-documented \citep{24, 25}. This approach is particularly adept at identifying collaborative behaviors manifest as transient correlation structures. However, the architectural focus on relational aggregation may naturally de-emphasize the fine-grained temporal evolution within individual channels.

The Mixture-of-Experts (MoE) paradigm \citep{26} has emerged as a promising architecture for integrating diverse modeling capabilities. Yet, its application in time series has largely been for scaling architecturally homogeneous models \citep{10} or for creating ensembles to tackle related but distinct problems like traffic flow prediction \citep{40}. This includes specialized MoE frameworks for tasks such as anomaly detection \citep{12, 28}, demonstrating the paradigm's flexibility, which builds on a long history of applying expert-based systems to complex classification problems \citep{33}. Despite the demonstrated power of heterogeneous MoE in other domains like multimodal learning \citep{29, 41} and graph learning \citep{27}, a unified framework that can dynamically arbitrate between these distinct analytical perspectives on a sample-by-sample basis remains an open research challenge.

\begin{figure*}
	\centering
	\includegraphics[width=0.85\linewidth]{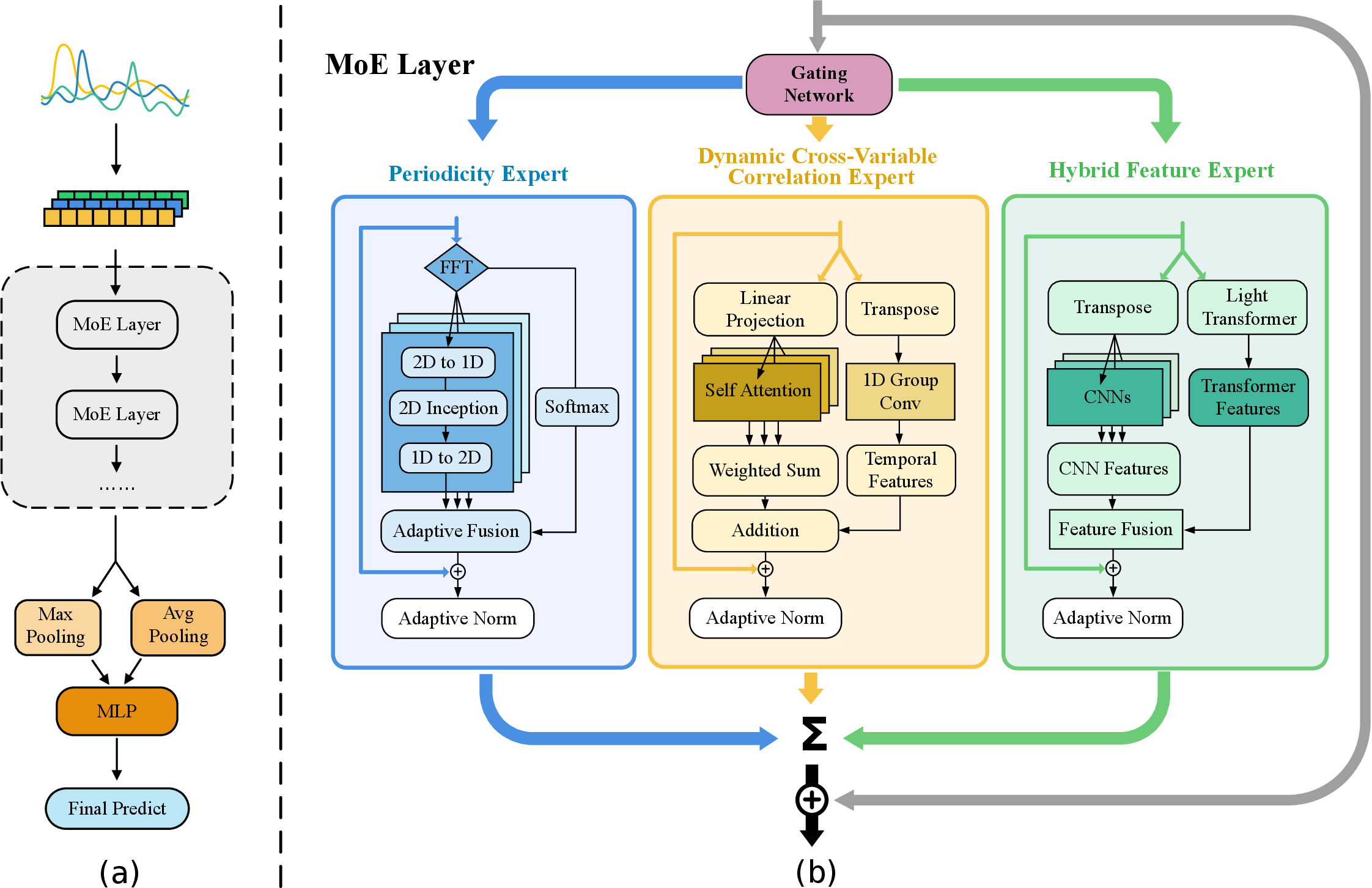}
	\caption{(a) Overall architecture of DAPNet, illustrating the flow of data through stacked MoE layers to a final prediction head. (b) Detailed structure of a single MoE layer, showing the gating network routing input to the three specialized experts whose outputs are adaptively fused.}
	\label{fig:2}
\end{figure*}

Collectively, while these specialized approaches have significantly advanced the field by mastering either temporal dynamics or cross-variable dependencies, they remain limited by their inherent architectural biases. While some efforts have sought to create monolithic hybrid models that fuse graph-based and sequential processing blocks \citep{49}, devising a unified framework that dynamically integrates these perspectives remains an open research challenge for network state classification.

\section{Method}
The task of network state classification is formally defined as a multivariate time-series classification problem. Given a dataset $\mathcal{D}=\{(X_i,y_i)\}_{i=1}^{N}$, where each sample $X_i \in \mathbb{R}^{T\times C}$ represents a segment of network activity over a time window of length $T$ with $C$ features (e.g., packet counts, flow duration), the corresponding label $y_i \in \{1,\dots,N_{\text{cls}}\}$ denotes one of $N_{\text{cls}}$ possible network states, which may include benign traffic and various types of attacks. The objective is to learn a mapping function $f\colon\mathbb{R}^{T\times C}\to\{1,\dots,N_{\text{cls}}\}$ that accurately predicts the state of a previously unseen sample.

For model processing, input samples are grouped into batches of size $B$. An initial embedding layer projects the $C$-dimensional features into a higher-dimensional space of size $d$. This yields the final input tensor for the network, denoted as $\mathbf{H}\in\mathbb{R}^{B\times T\times d}$.

\subsection{Model Architecture Overview}
DAPNet's core architecture is a hybrid mixture-of-experts model, as shown in \autoref{fig:2}. Its architecture is designed to move beyond the limitations of a single-model paradigm by acknowledging that different data samples may exhibit highly diverse intrinsic patterns. Therefore, DAPNet does not enforce a fixed processing workflow for all inputs. Instead, it operates on the principle of learning disentangled representations, where specialized experts are encouraged to capture distinct and complementary aspects of the data \citep{51}. It constructs a collection of multiple "expert" networks and uses a dynamic gating mechanism to adaptively select and weight the most appropriate experts for each input. Given the input feature tensor $\mathbf{H}$, the output $\mathbf{H}_{out}$ of the MoE layer is the weighted sum of the outputs from $N_e$ expert networks:

\begin{equation}
     \mathbf{H}_{out} = \sum_{i=1}^{N_e} g(\mathbf{H})_i \cdot E_i(\mathbf{H})
\end{equation}
where $g(\mathbf{H})_i$ is the scalar weight assigned by the gating network to the i-th expert, and $E_i(\mathbf{H})$ is the output tensor of the i-th expert. 

The gating network is a lightweight Multi-Layer Perceptron (MLP) that computes a routing score for each expert based on a global representation of the input sample, obtained via adaptive average pooling. To promote expert specialization and maintain computational efficiency, we employ a sparse routing strategy that activates only the top-$K$ experts with the highest scores. We set $K=2$ by default; this choice balances the need for specialization against the computational benefits of sparsity. The final weights $g(\mathbf{H})$ are obtained by applying a softmax function solely to the scores of the selected top-$K$ experts; all remaining experts receive zero weight.

\subsection{Expert Network Design}
The MoE layer contains three complementary expert networks, each designed to capture a unique aspect of time series data.

\subsubsection{Periodicity Expert} 
This expert adapts the core paradigm of TimesNet, specializing in distinguishing the complex, multi-scale periodicities of benign network operations from the arrhythmic disruptions or the newly imposed periodicities that characterize malicious activities. Benign network traffic possesses a distinct operational rhythm, a superposition of diverse cycles encompassing high-frequency protocol heartbeats and low-frequency administrative tasks, such as weekly backups. Malicious attacks fundamentally alter this rhythm, either by introducing chaotic, aperiodic traffic during a volumetric DDoS attack or by imposing a new, stealthy periodic signal, as seen in a botnet's command-and-control beaconing.

To formalize this, for an input time series $ \mathbf{H}\in \mathbb{R} ^{T \times d} $, the expert first computes its spectral representation via the Fast Fourier Transform (FFT) \citep{34} to create a spectral fingerprint of the network's state. 

\begin{equation}
 \hat{\mathbf{H}} = \text{FFT}(\mathbf{H}) 
\end{equation}

From the amplitudes $ \mid \hat{\mathbf{H}}\mid $, it identifies the top k dominant frequencies $ \left \{ f_1, \dots ,f_k \right \} $, which correspond to periods $ \left \{ P_1, \dots ,P_k \right \} $ where $ P_i = T/f_i $. For each identified period $ P_k $, the 1D time series is reshaped into a 2D tensor $ \mathbf{H}^{(k)}_{2D} \in \mathbb{R}^{{P_k} \times (T/P_k)} $. This transformation is not merely a formatting step; it reframes the analysis by explicitly separating intra-period variation (e.g., behavioral fluctuations within a single day) from inter-period variation (e.g., how daily patterns evolve from Monday to Friday). An Inception-style 2D CNN \citep{39} is then applied to each 2D representation to extract features, an approach that has proven highly effective for image-like representations of time series \citep{42}.

\begin{equation}
	F^{(k)} = \text{CNN2D}(H_{2D}^{(k)})
\end{equation}

By learning the spectral signature of normal operations, this expert can robustly identify traffic that deviates from the network's established tempo.

\subsubsection{Dynamic Cross-Variable Correlation Expert} 

\begin{figure}
	\centering
	\includegraphics[width=1\linewidth]{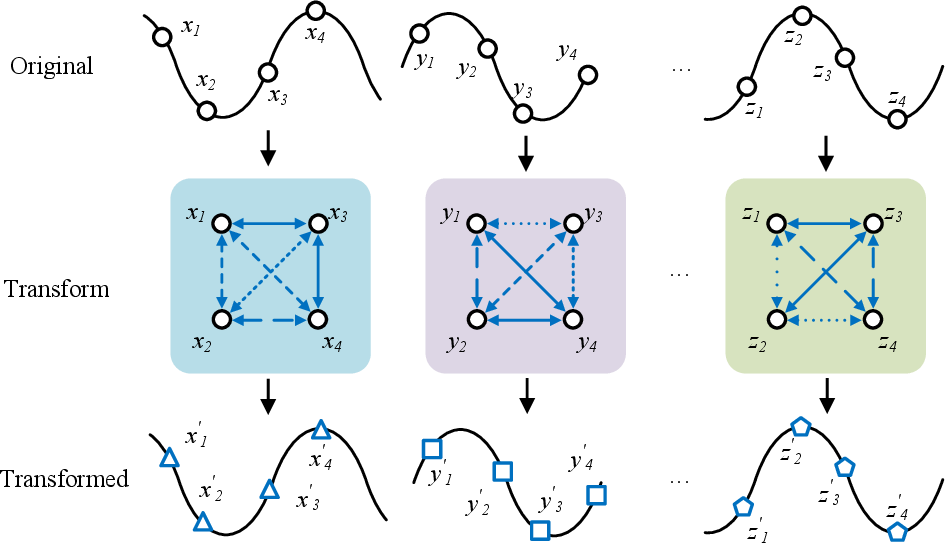}
	\caption{Illustration of the dynamic correlation mapping process.}
	\label{fig:3}
\end{figure}

This expert is designed to explicitly model the dynamic, time-varying dependencies among the C input features, which, in the context of network state classification, often represent distinct entities such as source/destination IP addresses, ports, or protocol types. As illustrated \autoref{fig:3}, the expert's core function is to transform a raw multivariate time series of network traffic ('Original') into a structured interaction graph ('Transformed') that reveals latent coordinated behaviors. Consequently, capturing these interdependencies is to model the coordinated interplay between these features during a network event. This allows the model to identify characteristic patterns of inter-variable correlation, such as the many-to-one convergence pattern of a Distributed Denial of Service (DDoS) attack or the one-to-many probing pattern of a port scan.

Operationally, the input tensor $ \mathbf{H} \in \mathbb{R} ^{B \times T \times d} $ is first transposed to $ \mathbb{R} ^{B \times d \times T} $ to prioritize the feature dimension. A self-attention mechanism is then applied across this dimension. The query (Q), key (K), and value (V) projections are computed from the transposed input. These are used to compute a sample-specific, weighted adjacency matrix $ A_{dyn} \in \mathbb{R} ^{d \times d} $ as follows:

\begin{equation}
 A_{dyn} = \text{Softmax}(\frac{QK^T}{\sqrt{d_k}})
\end{equation}
where $d_k$ is the dimension of the key and query vectors. 

Here, the matrix $A_{dyn}$ is the explicit materialization of the 'Transformed' graph in \autoref{fig:3}, where the value $ (A_{dyn})_{i,j} $ quantifies the learned influence of feature j on feature i. This allows the model to capture event-specific correlation structures, such as the many-to-one pattern of a DDoS attack or the one-to-many pattern of a port scan. Graph-based feature aggregation is then performed by multiplying this learned adjacency matrix with the value projection ($ \mathbf{H}_{agg} = A_{dyn}V $), allowing each feature's representation to be updated based on the state of its correlated peers. The final output combines this relational context with the original features via a residual connection.

\subsubsection{Hybrid Feature Expert}
This expert is designed to capture complex temporal patterns defined by an ordered sequence of events rather than by periodicity or cross-variable structure. This is particularly relevant for identifying threats like multi-stage APTs, which unfold over long time horizons, or obfuscated traffic, which may be characterized by local, high-frequency sequences of packet sizes and timings.

The expert employs a dual-branch architecture. The first branch, a local feature extractor composed of 1D convolutional layers with varying kernel sizes, captures short-term, high-frequency motifs. This design is based on the principles of Temporal Convolutional Networks (TCN) \citep{43}, which serves as an effective complement to the long-range dependency modeling of the Transformer branch:

\begin{equation}
 H_{CNN} = Concat(Conv1D_{k_1}(H), \dots, Conv1D_{k_n}(H)) 
\end{equation}

The second branch, a lightweight Transformer encoder, serves as a long-range dependency modeler. Its design is inspired by recent advances in efficient Transformers that reduce the canonical quadratic complexity to linear, for instance through the use of patching \citep{50}, enabling it to capture the relationships between events across the entire time window:

\begin{equation}
	H_{\text{Trans}} = \text{TransformerEncoder}(H)
\end{equation}

The features from both branches are concatenated and passed through a final linear layer, enabling the expert to analyze the sequential structure of network behavior at multiple temporal scales.

\begin{equation}
	H_{out} = Linear(Contact(H_{CNN}, H_{Trans}))
\end{equation}

\subsection{Hybrid Loss Function}
The training of DAPNet is guided by a dual-objective loss function to address two distinct challenges: an architectural challenge of maintaining the stability and functional diversity of the Mixture-of-Experts framework and a task-specific challenge of achieving high classification accuracy on network state datasets with severe class imbalance. Our hybrid loss function addresses these issues through two specialized components, a load-balancing loss and a classification loss, which are combined into a single optimization objective.

The load-balancing loss $L_{balance}$, is to preserve the architectural integrity of the MoE framework. This component is designed to prevent representational collapse, a documented training pathology in MoE models where the gating network routes most inputs to a small subset of experts. Such an outcome would render a subset of experts inactive, thereby negating the benefits of the multi-expert design. To counteract this, the load-balancing loss encourages an even distribution of inputs across the expert ensemble. This is a standard and essential technique for stable training of sparse MoE models, as demonstrated in foundational works like Switch Transformers \citep{44}, ensuring all experts are sufficiently utilized. It is defined as:

\begin{equation}
L_{balance} = N_e \cdot \sum_{i=1}^{N_e} f_i \cdot P_i
\end{equation}
where $N_e$ is the number of experts, $f_i$ is the fraction of samples in a batch routed to expert $i$, and $P_i$ is the average routing probability assigned to expert $i$ by the gating network.

Minimizing this term penalizes scenarios where a few experts dominate routing decisions, ensuring all experts remain active throughout the learning process.

The classification loss $L_{class}$, based on Focal Loss \citep{35}, is used to optimize classification performance, particularly for the imbalanced class distributions common in network intrusion detection. Standard cross-entropy loss can become biased towards the majority class, leading to poor detection of rare but critical attack types. Focal Loss mitigates this by introducing a modulating factor, $(1-p_t)^\gamma$, that dynamically down-weights the loss from well-classified, high-confidence examples. This forces the model to focus its capacity on more challenging, minority-class samples. The use of label smoothing further regularizes the model to prevent overconfident predictions and improve generalization.

The final training objective for DAPNet synthesizes these two components into a unified loss function:

\begin{equation}
     L_{total} = \alpha \cdot L_{balance} + \beta \cdot L_{class}
\end{equation}

\begin{equation}
	\delta = \frac{\alpha} \beta 
	\label{Eq:1}
\end{equation}

The hyperparameters $\alpha$ and $\beta$ provide direct control over the trade-off between the two competing objectives. The weight $\alpha$ governs the strength of the architectural regularization that ensures expert diversity, while $\beta$ scales the primary, task-specific objective of classification accuracy. The ratio $\delta$ between $\alpha$ and $\beta$ explicitly defines the balance between maximizing classification accuracy and ensuring all experts are trained. This provides a tunable and interpretable mechanism to adapt the training dynamics to the statistical properties of a given dataset, allowing for a principled approach to solving the dual challenges of MoE training and imbalanced classification.

\section{Experiments}
\subsection{Datasets}
To evaluate the performance and generalizability of DAPNet, the model was tested on two categories of datasets.

\begin{table*}[t]
\centering 
\resizebox{\textwidth}{!}{
\begin{tabular}{lccccccccc}
\toprule
\multirow{2}{*}{Data/Model} & MICN & Crossformer & LightTS & Transformer & TimesNet & TodyNet & ModernTCN & MPTSNet & \multirow{2}{*}{Ours} \\
& \small\textit{ICLR'23} & \small\textit{ICLR'23} & \small\textit{CoRL'22} & \small\textit{NeurIPS'17} & \small\textit{ICLR'23} & \small\textit{Info'24} & \small\textit{ICLR'24} & \small\textit{AAAI'25} & \\
\midrule
EthanolConcentration & 28.14 & 31.18 & 26.62 & 27.76 & 27.76 & \underline{35.00} & 30.80 & \textbf{39.54} & 31.18 \\
FaceDetection & 64.76 & 64.59 & 66.40 & 68.27 & 69.10 & 62.70 & 68.98 & \underline{69.55} & \textbf{70.20} \\
Handwriting & 8.71 & 10.12 & 6.94 & 28.47 & 28.82 & \underline{43.60} & 28.59 & 33.29 & \textbf{53.41} \\
Heartbeat & 73.17 & 75.61 & 76.10 & 75.61 & \underline{78.54} & 75.60 & 73.17 & 73.17 & \textbf{80.49} \\
Japanese Vowels & 70.27 & 95.41 & 75.41 & 95.14 & 97.03 & \underline{98.11} & 97.84 & 96.76 & \textbf{98.38} \\
PEMS-SF & 78.61 & 86.13 & 84.39 & 85.55 & 82.66 & 78.00 & 88.44 & \textbf{90.75} & \underline{88.44} \\
SelfRegulationSCP1 & 87.37 & \underline{91.81} & 89.76 & 90.10 & 90.10 & 89.80 & 90.78 & \textbf{93.17} & 91.13 \\
SelfRegulationSCP2 & 53.89 & 52.22 & 55.56 & 54.44 & 54.44 & 55.00 & \underline{56.67} & 53.89 & \textbf{58.33} \\
SpokenArabicDigits & 96.82 & 98.00 & 98.04 & 98.27 & 98.91 & 97.37 & 98.14 & \underline{99.00} & \textbf{99.82} \\
UWaveGestureLibrary & 78.75 & 79.38 & 78.75 & 86.56 & 86.56 & 85.00 & 85.94 & \underline{87.19} & \textbf{90.10} \\
\midrule 
CICIDS2017 & 97.88 & 83.11 & \underline{97.92} & 93.34 & 97.27 & 96.03 & 97.19 & 96.66 & \textbf{99.50} \\
CICIDS2018 & 99.44 & 77.50 & 99.10 & 98.84 & 99.51 & 98.79 & \underline{99.61} & 99.42 & \textbf{99.70} \\
\bottomrule
\end{tabular}
} 
\caption{Comparison of accuracy on ten UEA datasets and two CICIDS datasets. The highest result is indicated in bold, and the second highest result is indicated with an underline.}
\label{tab:1}
\end{table*}

\textbf{UEA benchmark datasets} We employed 10 widely-used datasets from the UEA Multivariate Time Series Classification Archive \citep{17}. These datasets span various domains (e.g., motion, speech, sensor signals) and vary in sequence length, number of variables, and class categories, providing a thorough benchmark for assessing the model’s generalization ability.

\textbf{CICIDS network security datasets} To evaluate DAPNet's performance in its target application domain, we used two large-scale network intrusion detection datasets: CICIDS2017 and CICIDS2018 \citep{18}. These datasets contain millions of network traffic records and various attack types, presenting challenges such as extreme class imbalance and high dimensionality. For efficient training and evaluation, we randomly sampled 500,000 instances from each dataset using stratified sampling to preserve the original class distributions.

\subsection{Baselines}
To ensure a fair and comprehensive comparison, we selected representative models from different technical paradigms as baselines. These include: MPTSNet \citep{7}, TodyNet \citep{8}, ModernTCN \citep{15}, Crossformer \citep{13}, MICN \citep{16}, LightTS \citep{14}, TimesNet \citep{6}, Transformer \citep{3}. 

\subsection{Evaluation Metrics}
The main evaluation metric is classification accuracy. Considering the severe class imbalance in the network intrusion datasets, we additionally report precision, recall, and F1-score for each attack category in our ablation experiments to provide a more detailed and balanced performance evaluation.

\subsection{Implementation Details}
All experiments were conducted within the PyTorch framework, utilizing a single NVIDIA GeForce 4060 Ti GPU for acceleration. The model was trained using the AdamW optimizer with an initial learning rate of $10^{-3}$, which was managed by a cosine annealing schedule with a warm-up phase to ensure stable convergence. To prevent overfitting, an early stopping mechanism was employed with a patience of 10 epochs, halting the training process if the validation loss did not improve. The hidden feature dimension, denoted as $d\_model$, was selected from the set {64, 128, 256}, with the specific choice depending on the scale and complexity of the individual dataset. The Mixture-of-Experts (MoE) layer, a core component of the architecture, was configured with three distinct experts ($N_e$=3) by default. A sparse routing strategy was implemented, activating only the top two experts for any given input sample to maintain computational efficiency. 

The impact of the balance coefficient $\delta$ in the hybrid loss function on performance and its relationship with dataset complexity will be analyzed in detail in section \ref{sec:1}.

\begin{figure*}
    \centering
    \includegraphics[width=1\linewidth]{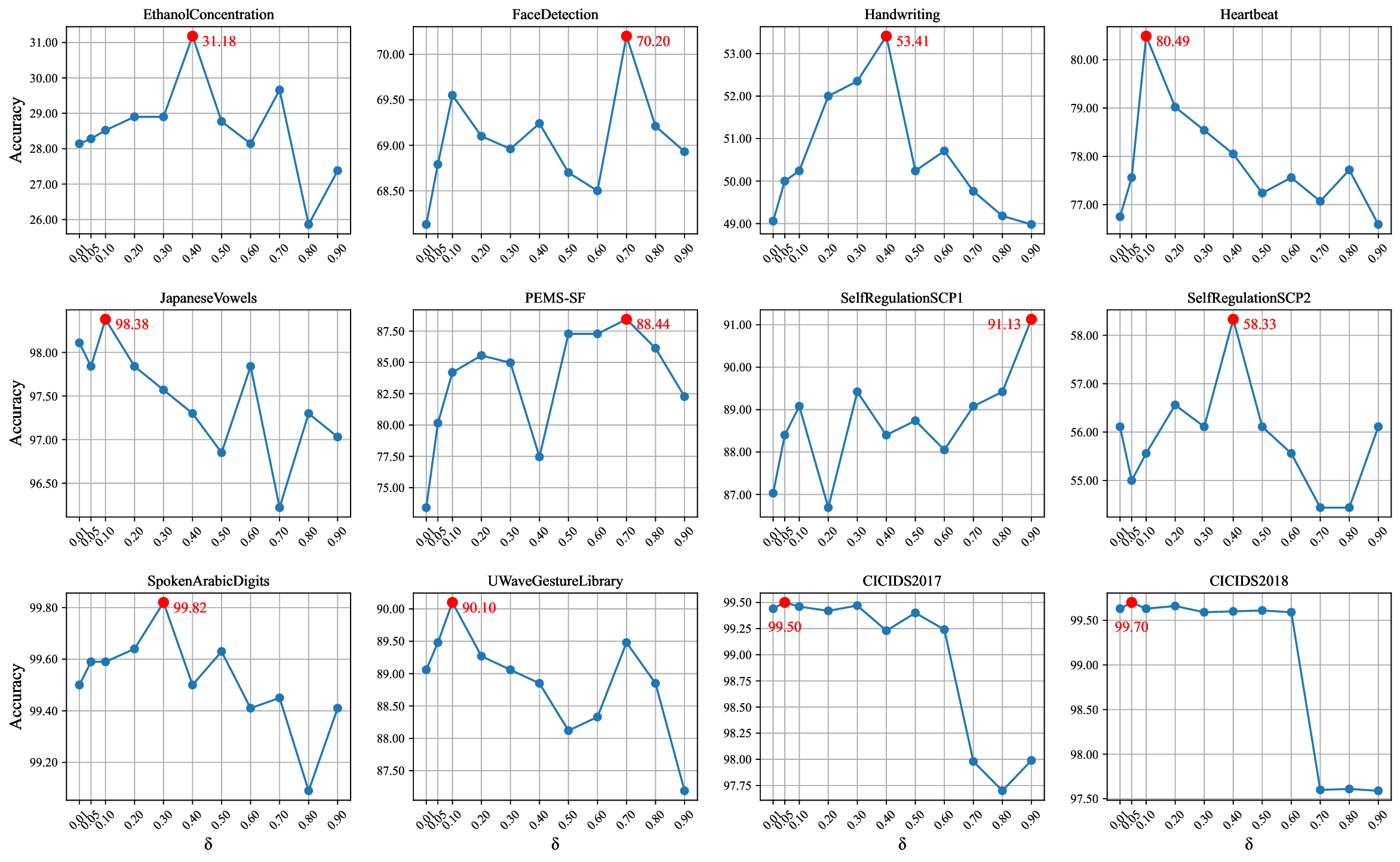}
    \caption{The impact of the balance coefficient $\delta$ in the hybrid loss function on accuracy under different levels of data complexity.}
    \label{fig:4}
\end{figure*}

\subsection{Experimental Results}

\subsubsection{Overall Model Performance}
As shown in \autoref{tab:1}, DAPNet achieved the highest accuracy on 7 out of the 10 evaluated UEA datasets. The design of adaptive Mixture-of-Experts (MoE) architecture allows it to dynamically deploy the most suitable analytical approach for each dataset's unique characteristics. For instance, the Periodicity Expert is crucial for isolating faint, cyclical signals in low signal-to-noise ratio datasets like Heartbeat. Conversely, the Dynamic Cross-Variable Correlation Expert provides a distinct advantage on datasets with strong inter-variable dependencies, such as PEMS-SF. Its ability to learn a sample-specific correlation matrix allows it to effectively capture the transient, collaborative patterns among the numerous sensors in the dataset. For tasks requiring fine-grained shape recognition, like Handwriting, the Hybrid Feature Expert excels by integrating local and global temporal features. 

It is plausible that for datasets like EthanolConcentration and PEMS-SF, the classification task is overwhelmingly dominated by these frequency-based, periodic features. In such cases, a highly specialized model like MPTSNet, which dedicates its entire architecture to this single analytical paradigm, may achieve superior performance.

On the CICIDS2017 and CICIDS2018 datasets, DAPNet achieves accuracies of 99.50\% and 99.70\% respectively, outperforming the evaluated baseline models. The model's performance on these datasets can be attributed to its architectural design, which addresses domain-specific challenges. The task is less about subtle periodicities and more about robust feature extraction in a high-dimensional space, a task well-suited to the powerful Hybrid Feature Expert. Most critically, the Focal Loss component of the hybrid loss function is designed to address the class imbalance common in security datasets. By down-weighting the loss from easily classified benign samples, it forces the model to focus on learning the signatures of rare but critical attack classes, which is essential for reliable threat detection.

\subsubsection{Impact of $\delta$ on Accuracy}
\label{sec:1}
The hyperparameter $\delta$ in DAPNet's hybrid loss function, defined in Eq. \eqref{Eq:1}, plays a critical role in mediating the trade-off between the primary classification objective and the stability of the Mixture-of-Experts (MoE) training process. We observed an interesting correlation between the optimal $\delta$ and the characteristics of a given dataset. We hypothesize that datasets with greater pattern heterogeneity may require a stronger balancing force (a higher $\delta$) to ensure all specialized experts are effectively utilized. The analysis in \autoref{fig:4} explores this relationship, investigating how model accuracy fluctuates with different values of $\delta$ across various datasets. This offers an empirical insight into how the balance between task performance and expert utilization shifts according to the perceived pattern diversity in the data.

Datasets with low complexity, such as SpokenArabicDigits and JapaneseVowels, exhibit homogeneous patterns and achieve peak performance with a small $\delta$ (around 0.1–0.2). This same principle applies to the CICIDS2017 and CICIDS2018 network traffic datasets. Although large and seemingly disorganized, their attacks are repetitive and generated by fixed tools, which also results in low complexity. Consequently, for all such datasets, a single expert is sufficient, and a strong balancing loss proves counterproductive.

Datasets of medium complexity, like Heartbeat and SelfRegulationSCP2, present mixed challenges like subtle signals within noise and thus benefit from a moderate $\delta$ (around 0.5–0.6) that encourages collaboration between specialist and generalist experts. 

Datasets with high complexity, including PEMS-SF, Handwriting, and FaceDetection, are characterized by high pattern heterogeneity. These tasks require a large $\delta$ (0.7–0.9) to enforce exploration and ensure all experts are trained collaboratively to model the diverse, orthogonal patterns present in the data. This model-centric view offers a more insightful measure of complexity and validates the need for adaptive architectures.

\begin{figure}
    \centering
    \includegraphics[width=1\linewidth]{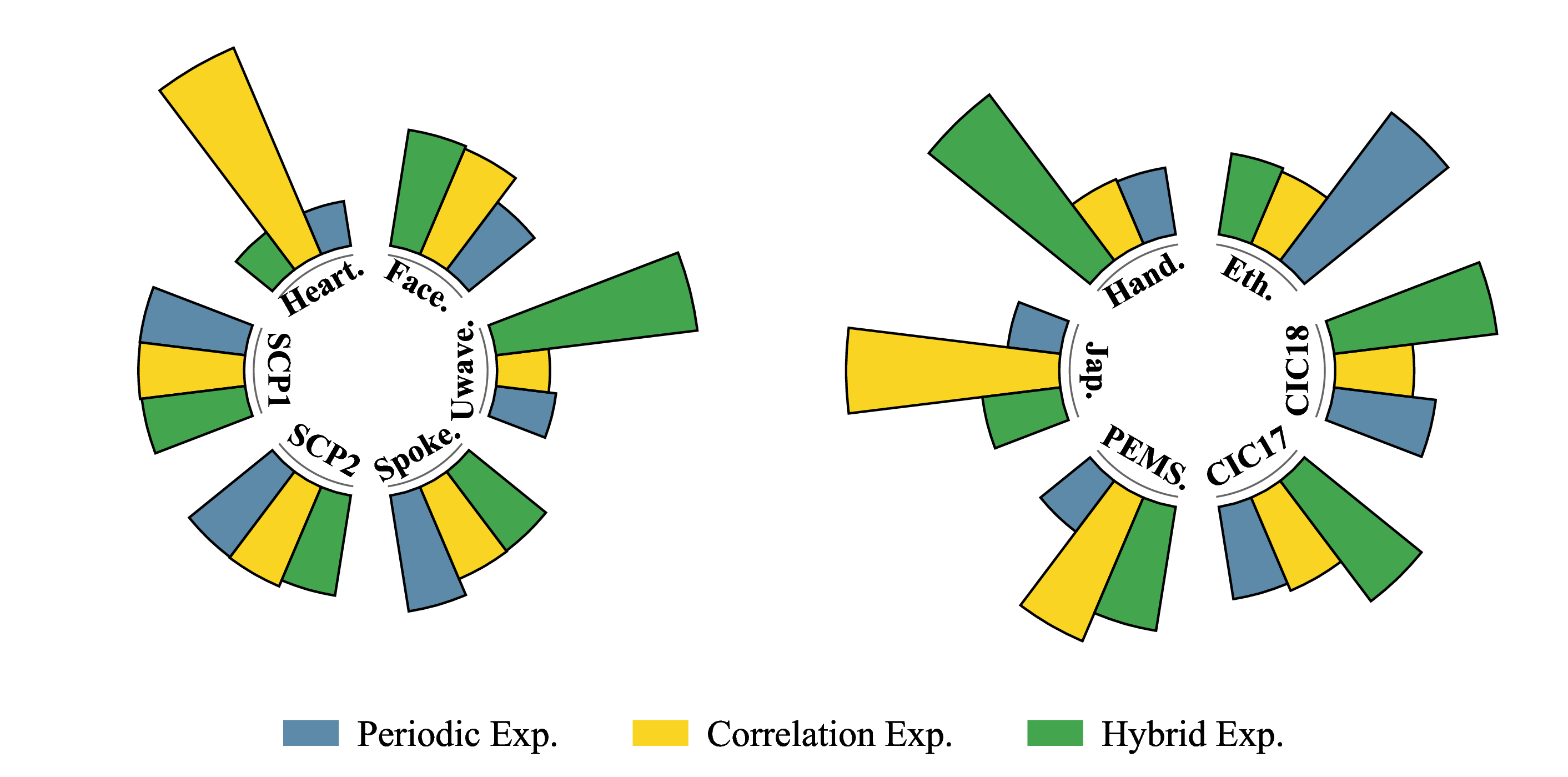}
    \captionof{figure}{Visualization of DAPNet's adaptive expert routing mechanism.}
    \label{fig:5}
\end{figure}

\subsubsection{Analysis of Adaptive Expert Routing}
\autoref{fig:5} provides empirical evidence for the effectiveness of DAPNet's adaptive routing mechanism, demonstrating that the model learns a meaningful, data-driven routing strategy instead of acting as a static ensemble. The gating network's behavior correctly assigns higher weights to the Dynamic Cross-Variable Correlation Expert on datasets where inter-variable dependency is the dominant feature, such as the high-dimensional PEMS-SF and FaceDetection datasets. Similarly, for data like Heartbeat, where the main challenge is extracting periodic signals from noise, the Periodicity Expert receives a greater allocation of resources. On datasets with simpler, more clearly defined patterns like UWaveGestureLibrary, the versatile Hybrid Feature Expert is often preferred. This alignment between dataset characteristics and expert selection suggests that the adaptive routing mechanism contributes to DAPNet's performance.

\begin{table*}[t]
\centering
\resizebox{\textwidth}{!}{
\begin{tabular}{lcccccccccccc}
\toprule
Model/Data & Eth. & Face. & Hand. & Heart. & Jap. & PEMS. & SCP1 & SCP2 & Spoke. & Uwave. & CIC.17 & CIC.18 \\
\midrule
w/o Expert 2,3 & 28.06 & 68.21 & 27.34 & 74.34 & 98.00 & 87.17 & 87.99 & 54.44 & 99.64 & 85.25 & 98.05 & 99.62  \\
w/o Expert 1,2 & 27.76 & 66.48 & 27.46 & 75.22 & 98.22 & \underline{87.51} & 88.33 & 53.67 & 99.66 & 83.63 & 99.44 & \underline{99.72}  \\
w/o Expert 1,3 & 29.05 & 67.37 & 27.39 & 76.00 & 98.32 & 84.05 & 89.28 & 55.67 & \textbf{99.83} & 80.25 & 96.77 & 99.43  \\
w/o MoE Routing & 28.14 & 68.48 & 28.66 & 75.90 & \textbf{98.43} & 86.47 & \underline{90.72} & \underline{56.00} & 99.54 & 83.19 & \textbf{99.53} & \textbf{99.87}  \\
w/ CE Loss & \underline{29.58} & \underline{68.88} & \underline{49.72} & \underline{78.44} & 97.46 & 83.01 & 88.67 & 55.67 & 99.65 & \underline{89.44} & 99.31 & 99.62 \\
\midrule
DAPNet & \textbf{31.18} & \textbf{70.20} & \textbf{53.41} & \textbf{80.49} & \underline{98.38} & \textbf{88.44} & \textbf{91.13} & \textbf{58.33} & \underline{99.82} & \textbf{90.10} & \underline{99.50} & 99.70 \\
\bottomrule
\end{tabular}
}
\caption{Average accuracy (\%) of ablation experiments on 12 datasets using DAPNet. The highest result is indicated in bold, and the second highest result is indicated with an underline. Expert 1: Periodicity, Expert 2: Dynamic Cross-Variable Correlation, Expert 3: Hybrid} 
\label{tab:2}
\end{table*}

\begin{figure}
	\centering
	\includegraphics[width=1\linewidth]{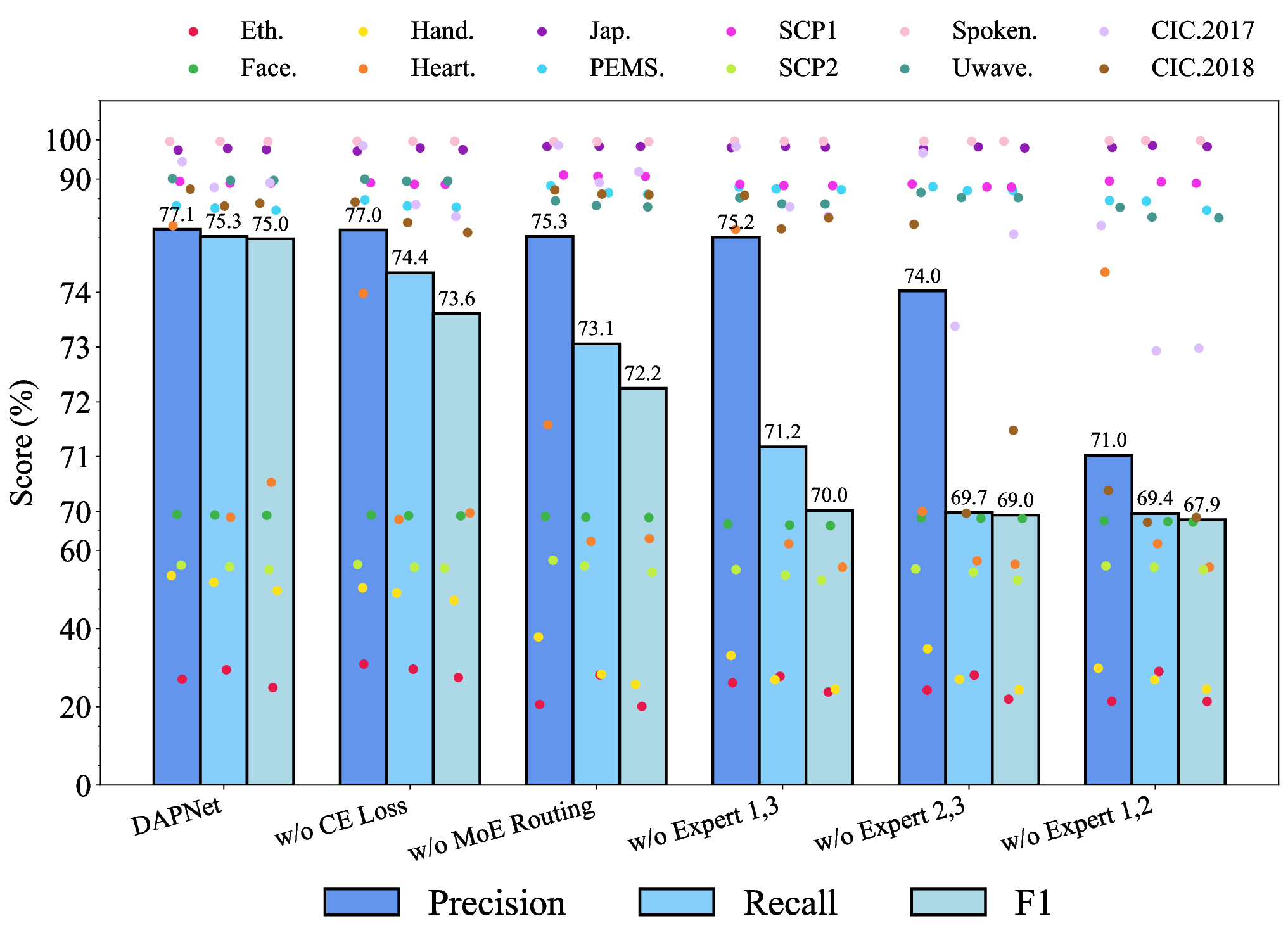}
	\caption{The comparison of precision, recall, and F1 scores of each ablation variant across all datasets demonstrates the performance degradation relative to the complete DAPNet model. Each different colored dot represents the model's parameters on a specific dataset.}
	\label{fig:6}
\end{figure}

\subsection{Ablation Studies}
To thoroughly evaluate the effectiveness of each key component in DAPNet, we conducted extensive ablation experiments. To eliminate the random effects of chance and ensure the stability and reliability of the results, all performance metrics reported in this section are the average values obtained after running the model independently under five different random seeds (42, 123, 456, 789, 2025). Based on the full DAPNet architecture, we constructed the following five main variants:

\begin{itemize}[leftmargin=*]
\item \textbf{w/o Expert 2,3}: Remove the Dynamic Cross-Variable Correlation Expert (Expert 2) and Hybrid Feature Expert (Expert 3), retaining only the Periodicity Expert (Expert 1). Tests the independent effectiveness of the periodicity module.

\item \textbf{w/o Expert 1,2}: Retains only the Hybrid Feature Expert.

\item \textbf{w/o Expert 1,3}: Retains only the Dynamic Cross-Variable Correlation Expert.

\item \textbf{w/o MoE Routing}: Replace the dynamic gated routing mechanism with simple average ensemble (equal weights for all experts). Tests whether performance gain arises from intelligent routing.

\item \textbf{w/ CE Loss}: Replaces the hybrid loss (Focal Loss + load balancing) with standard cross-entropy loss, to validate the loss function’s effectiveness in handling class imbalance and optimizing the MoE architecture.
\end{itemize}

The detailed results of the ablation experiments are shown in \autoref{tab:2} and \autoref{fig:6}, from which the following conclusions can be drawn:

The MoE dynamic routing mechanism is crucial. When we replaced the dynamic MoE gated routing mechanism with simple average ensemble (i.e., the mean of all expert outputs), the F1 score decreased by 2.73\%. This demonstrates that the core advantage of DAPNet does not stem from simple model ensemble but from its ability to intelligently and dynamically route based on the characteristics of the input data.

Each expert is indispensable. Removing any one of the three expert networks results in a significant decline in model performance. This result indicates that the three experts are functionally complementary and non-redundant, each capturing information from different dimensions to collectively form DAPNet's powerful representational capability.

The effectiveness of the hybrid loss function. Replacing our designed hybrid loss function (Focal Loss + load balancing loss) with the standard cross-entropy loss (CE Loss) resulted in a 1.37\% decline in F1 score. This confirms that our loss function design effectively addresses class imbalance issues and ensures stable and adequate training of the MoE architecture through load balancing loss, thereby enhancing overall performance.

\section{Conclusion}
This paper tackles two fundamental challenges in network state classification: complex temporal periodicity and dynamic inter-variable correlation. To address these, we introduce DAPNet—a novel Dynamic Adaptive Parsing Network purpose-built to integrate the three analytical paradigms most relevant to this domain: periodic analysis, dynamic cross-variable correlation modeling, and hybrid temporal feature extraction. Through an innovative Mixture-of-Experts (MoE) architecture combined with an intelligent gating mechanism, DAPNet adaptively selects suitable analytical pathways tailored to diverse data characteristics. Extensive experiments demonstrate the competitive performance and generalization capability of DAPNet compared to existing state-of-the-art models.

Our findings support a central hypothesis: that a robust solution to network state classification demands a framework that can dynamically arbitrate between the domain's core analytical dimensions. By demonstrating competitive performance on security datasets, DAPNet is presented as a promising framework for network state classification.

\section{Limitations and Future Work}
While DAPNet achieves competitive performance on the evaluated datasets, several limitations are acknowledged. The core trade-off of this architecture lies in computational complexity. Although efficient inference is achieved through sparse routing, the resource consumption of training three independent experts is still higher than that of a single-paradigm model. The model is sensitive to the hyperparameter $\delta$ and requires dataset-specific tuning to balance classification accuracy and expert utilization. Finally, although gated weights provide some insights, the experts themselves still operate largely as a "black box" and require further research to improve interpretability. 

Based on DAPNet's success, several promising research directions emerge. Immediate work includes developing more lightweight experts to reduce training complexity and extending the framework to pretraining and continual learning for adapting to evolving threats. More advanced explorations could enhance the Dynamic Cross-Variable Correlation Expert's temporal granularity by modeling intra-window dynamics, rather than a single graph per sample.

\section{Acknowledgments}
This research was funded by the Fundamental Research Funds for the Central Universities(Grant Number 3282024052, 3282024058).

\bibliographystyle{cas-model2-names}
\bibliography{cas-refs}

\end{document}